\documentclass[10pt,a4paper]{llncs}
\usepackage[british]{babel} 
\usepackage[utf8]{inputenc}
\usepackage[T1]{fontenc}
\usepackage[misc]{ifsym}
\usepackage{amsmath,cmap,graphicx,multirow,algorithm,algpseudocode,txfonts,units,enumitem,tabularx,microtype,booktabs,xspace}
\usepackage{fancyhdr}

\DeclareMathOperator{\argmin}{argmin} % argmin

\newcommand{\gC}{\ensuremath{C}} % number of identity classes
\newcommand{\gD}{\ensuremath{D}} % dimensionality
\newcommand{\gDELTA}{\ensuremath{\gB{\Delta}}} % diagonal matrix
\newcommand{\gDELTAccH}[2]{\ensuremath{\gH{\delta}\left(#1,#2\right)}} % distance function in feature space
\newcommand{\gG}{\ensuremath{\mathcal{G}}} % space
\newcommand{\gGn}[1]{\ensuremath{\gB{g}_{#1}}} % gait pattern
\newcommand{\gGnH}[1]{\ensuremath{\gH{\gB{g}}_{#1}}} % gait pattern in feature space
\newcommand{\gGnc}[2]{\ensuremath{\gGn{#1}^{(#2)}}} % gait pattern by category
\newcommand{\gGAMMAjt}[2]{\ensuremath{\gamma_{#1}\left(#2\right)}} % coordinates
\newcommand{\gIc}[1]{\ensuremath{\mathcal{I}_{#1}}} % identity class
\newcommand{\gJ}{\ensuremath{J}} % number of joints
\newcommand{\gLAMBDAn}[1]{\ensuremath{\ell_{#1}}} % label of identity class
\newcommand{\gM}{\ensuremath{\mu}} % mean of whole data set
\newcommand{\gMc}[1]{\ensuremath{\gM_{#1}}} % mean of identity class
\newcommand{\gMcH}[1]{\ensuremath{\gH{\gM}_{#1}}} % mean of identity class in feature space
\newcommand{\gN}{\ensuremath{N}} % number of gait cycles
\newcommand{\gNc}[1]{\ensuremath{\gN_{#1}}} % number of gait cycles in category
\newcommand{\gPHI}{\ensuremath{\gB{\Phi}}} % transformation matrix of selected eigenvalues
\newcommand{\gPSI}{\ensuremath{\gB{\Psi}}} % transformation matrix of all eigenvalues
\newcommand{\gSIGMAc}[1]{\ensuremath{\gB{\Sigma}_{#1}}} % scatter matrix of identity class
\newcommand{\gSIGMAb}{\ensuremath{\gSIGMAc{\mathrm{B}}}} % between-class scatter matrix
\newcommand{\gSIGMAw}{\ensuremath{\gSIGMAc{\mathrm{W}}}} % within-class scatter matrix
\newcommand{\gSIGMAt}{\ensuremath{\gSIGMAc{\mathrm{T}}}} % total scatter matrix
\newcommand{\gT}{\ensuremath{T}} % average length of gait cycles

\newcommand{\gTHETA}{\ensuremath{\gB{\Theta}}}
\newcommand{\gXI}{\ensuremath{\gB{\Xi}}}
\newcommand{\gUPSILON}{\ensuremath{\gB{\Upsilon}}}

\newcommand{\gCHI}{\ensuremath{\gB{X}}}

\newcommand{\gOMEGA}{\ensuremath{\gB{\Omega}}}

\newcommand{\gB}[1]{\ensuremath{\mathbf{#1}}} % corresponding symbol in bold
\newcommand{\gH}[1]{\ensuremath{\widehat{#1}}} % corresponding symbol in feature space
\newcommand{\gO}[1]{\ensuremath{\overline{#1}}} % corresponding symbol in PCA-given space
\newcommand{\gL}[1]{\ensuremath{{#1}_L}} % corresponding symbol in learning part
\newcommand{\gE}[1]{\ensuremath{{#1}_E}} % corresponding symbol in evaluation part

\newcommand{\sub}[1]{\raisebox{-.4ex}{\scriptsize{#1}}}
\newcommand\etal{\textit{et al.}\xspace}
\bgroup
\catcode`_=\active
\gdef\underworks{\catcode`_=\active
\def_{\setbox0=\hbox{0}\hskip\wd0\relax}}
\egroup

\usepackage[breaklinks=true,colorlinks,bookmarks=false]{hyperref}

\begin{document}

\title{An~Evaluation~Framework~and~Database for~MoCap-Based~Gait~Recognition~Methods\thanks{This is a companion paper to our papers~\cite{BS16a,BS16b}.}}
\author{Michal Balazia (\href{https://orcid.org/0000-0001-7153-9984}{0000-0001-7153-9984}) \and Petr Sojka (\href{https://orcid.org/0000-0002-5768-4007}{0000-0002-5768-4007})}

\institute{Faculty of Informatics, Masaryk University, Botanick\'a 68a, 602\,00 Brno, Czech Republic\\{\tt xbalazia@mail.muni.cz} and {\tt sojka@fi.muni.cz}}

\maketitle

\pagestyle{plain}
\thispagestyle{fancy}
\fancyhead[C]{\ifthenelse{\value{page}=1}{1st Workshop on Reproducible Research in Pattern Recognition 2016, preprint}}
\headheight23pt

\begin{abstract}
As a contribution to reproducible research, this paper presents a framework and a database to improve the development, evaluation and comparison of methods for gait recognition from motion capture (MoCap) data. The evaluation framework comprises source codes of state-of-the-art human-interpretable geometric features as well as our own approaches where gait features are learned by a~modification of Fisher's Linear Discriminant Analysis with the Maximum Margin Criterion, and by a combination of Principal Component Analysis and Linear Discriminant Analysis. It includes a description and source codes of a~mechanism for evaluating class separability coefficients of feature space and four classifier performance metrics. This framework also contains a tool for learning a custom classifier and for classifying a custom probe on a custom gallery. We provide an experimental database along with source codes for its extraction from the general CMU MoCap database.
\end{abstract}

\section{Introduction}
\label{intro}
Gait (walk) pattern has several attractive properties as a soft biometric trait. From a~surveillance perspective, gait pattern biometrics is appealing in that it can be performed at a distance without requiring body-invasive equipment or subject cooperation.

Many research groups investigate the discrimination power of gait pattern and develop models that are applied to the automatic recognition of walking people from MoCap data. A number of MoCap-based gait recognition methods have been introduced in the past few years and new ones continue to emerge. In order to move forward with this competitive research, it is necessary to compare their innovative approaches with the state-of-the-art and evaluate them against established evaluation metrics on a benchmark database. New frameworks and databases have been developed recently~\cite{CK13,KTTEF15}.

As a contribution to reproducible research, this paper focuses on our framework for evaluating MoCap-based gait recognition methods and our benchmark MoCap gait database. We provide a large experimental database together with its extraction-and-normalization drive from the general CMU MoCap database, as specified in Section~\ref{db}. Implementation details of thirteen relevant methods are summarized in Section~\ref{impl}. In Section~\ref{eval} we describe the evaluation mechanism and define four class separability coefficients and four rank-based classifier performance metrics. Finally, Section~\ref{repro} consists of a manual and comments on reproducing the experiments.

\section{Data}
\label{db}

\begin{figure}[ht]
\vspace{-8pt}
\centering
\includegraphics[width=0.7\textwidth]{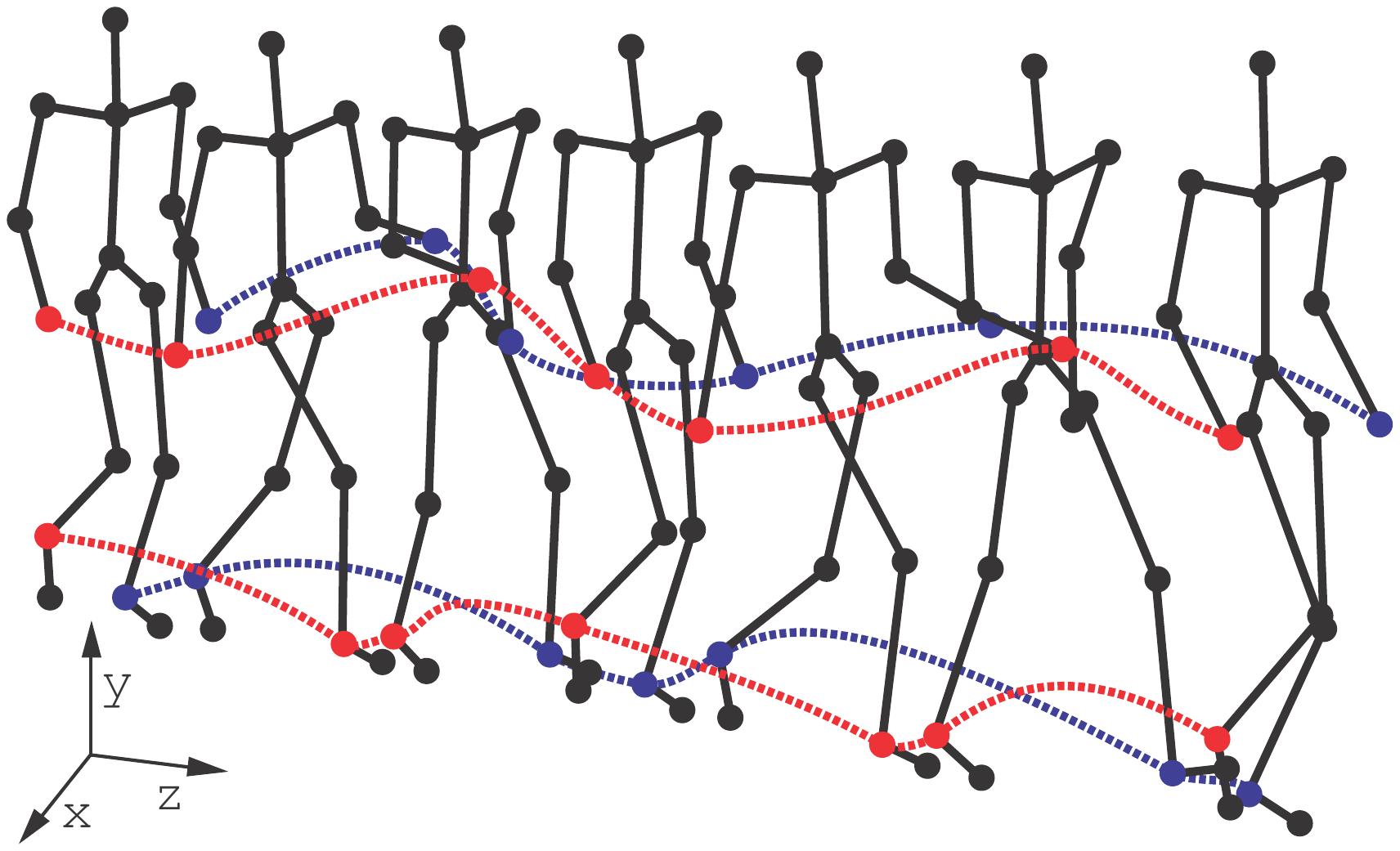}
\vspace{-5pt}
\caption{Motion capture data. A skeleton is represented by a stick figure of 31~joints (only 17 are shown here). Seven selected video frames of a walk sequence contain 3D coordinates of each joint in time. The red and blue lines track trajectories of hands and feet.~\cite{VBZ16}}
\label{f1}
\vspace{-5pt}
\end{figure}

MoCap technology provides video clips of individuals walking which contain structural motion data. The format keeps an overall structure of the human body and holds estimated 3D positions of major anatomical landmarks as the person moves. These MoCap data can be collected online by a system of multiple cameras (Vicon) or a depth camera (Microsoft Kinect). To visualize MoCap data (see Figure~\ref{f1}), a simplified stick figure representing the human skeleton (graph of joints connected by bones) can be recovered from body point spatial coordinates in time. Recent rapid improvement in MoCap sensor accuracy has brought affordable MoCap technology to assist human identification in such applications as access control and video surveillance.

For evaluation purposes we have extracted a large number of gait samples from the MoCap database obtained from the CMU Graphics Lab~\cite{CMU03}, which is available under the Creative Commons license. It is a well-known and recognized database of structural human motion data and contains a considerable number of gait sequences. Motions are recorded with an optical marker-based Vicon system. People wear a black jumpsuit with 41~markers taped on. The tracking space of \unit[30]{m$^2$} is surrounded by 12~cameras with a~sampling rate of \unit[120]{Hz} at heights ranging from 2 to 4~meters above ground thereby creating a video surveillance environment. Motion videos are triangulated to get highly accurate 3D data in the form of relative body point coordinates (with respect to the root joint) in each video frame and are stored in the standard ASF/AMC data format. Each registered participant is assigned with their respective skeleton described in an ASF file. Motions in the AMC files store bone rotational data, which is interpreted as instructions about how the associated skeleton deforms over time.

These MoCap data, however, contain skeleton parameters pre-calibrated by the CMU staff. Skeletons are unique to each walker and even a trivial skeleton check could result in \unit[100]{\%}~recognition. In order to fairly use the collected data, a prototypical skeleton is constructed and used to represent bodies of all subjects, shrouding the skeleton parameters. Assuming that all walking individuals are physically identical disables the skeleton check from being a potentially unfair classifier. Moreover, this is a skeleton-robust solution as all bone rotational data are linked to one specific skeleton. To obtain realistic parameters, it is calculated as mean of all skeletons in the provided ASF files.

The raw data are in the form of bone rotations or, if combined with the prototypical skeleton, 3D joint coordinates. The bone rotational data are taken from the AMC files without any pre-processing. We calculate the joint coordinates using the bone rotational data and the prototypical skeleton. One cannot directly use raw values of joint coordinates, as they refer to absolute positions in the tracking space, and not all potential methods are invariant to person's position or walk direction. To ensure such invariance, the center of the coordinate system is moved to the position of root joint $\gGAMMAjt{\mathrm{root}}{t}=[0,0,0]^\top$ for each time~$t$ and the axes are adjusted to the walker's perspective: the X~axis is from right (negative) to left (positive), the Y~axis is from down (negative) to up (positive), and the Z~axis is from back (negative) to front (positive). In the AMC file structure notation it is achieved by setting the root translation and rotation to zero (\texttt{root 0 0 0 0 0 0}) in all frames of all motion sequences.

Since the general motion database contains all motion types, we extracted a number of sub-motions that represent gait cycles. First, an exemplary gait cycle was identified, and clean gait cycles were then filtered out using a threshold for their Dynamic Time Warping (DTW) distance on bone rotations in time. The distance threshold was explicitly set low enough so that even the least similar sub-motions still semantically represent gait cycles. Setting this threshold higher might also qualify sub-motions that do not resemble gait cycles anymore. Finally, subjects that contributed with less than 10~samples were excluded. The final database~\cite{WWW2} has 54~walking subjects that performed 3,843~samples in total, which results in an average of about 71~samples per subject.
 
\section{Implementation Details of Algorithms}
\label{impl}

Recognizing a person from their gait involves capturing and normalizing their walk sample $\gGn{n}$, extracting gait features to compose a template $\gGnH{n}$, and finally querying the gallery for a~set of similar templates $\gGnH{n'}$~--~based on a distance function $\gDELTAccH{\gGnH{n}}{\gGnH{n'}}$~--~to report the most likely identity. This work focuses on extracting robust and discriminative gait features from raw MoCap data.

Many geometric gait features have been introduced over the past few years. They are typically combinations of static body parameters (bone lengths, person's height)~\cite{KKMJ14} with dynamic gait features such as step length, walking speed, joint angles and inter-joint distances~\cite{APG15,AA15,KKMJ14,RCA15}, along with various statistics (mean, standard deviation or maximum) of their signals~\cite{AWLSWZ16}. Clearly, these features are schematic and human-interpretable, which is convenient for visualizations and for intuitive understanding, but unnecessary for automatic gait recognition. Instead, our approach~\cite{BS16a,BS16b} prefers learning features in a~supervised manner that maximally separate the identity classes and are not limited by such dispensable factors.

What follows is a detailed specification of the thirteen gait features extraction methods that we have reviewed in our work to date. Since the idea behind each method has some potential, we have implemented each of them for direct comparison.

\def\method #1 {\item[\textbullet] \textbf{#1}\space}
\begin{itemize}
\method Ahmed by Ahmed~\etal~\cite{AAS14} extracts the mean, standard deviation and skew during one gait cycle of horizontal distances (projected on the Z~axis) between feet, knees, wrists and shoulders, and mean and standard deviation during one gait cycle of vertical distances (Y~coordinates) of head, wrists, shoulders, knees and feet, and finally the mean area during one gait cycle of the triangle of root and two feet.
\method Ali by Ali~\etal~\cite{AWLSWZ16} measures the mean areas during one gait cycle of lower limb triangles.
\method Andersson by Andersson~\etal~\cite{AA15} calculates gait attributes as mean and standard deviation during one gait cycle of local extremes of the signals of lower body angles, step length as a maximum of feet distance, stride length as a length of two steps, cycle time and velocity as a ratio of stride length and cycle time. In addition, they extract the mean and standard deviation during one gait cycle of each bone length, and height as the sum of the bone lengths between head and root plus the averages of the bone lengths between root and both feet.
\method Ball by Ball~\etal~\cite{BRRV12} measures mean, standard deviation and maximum during one gait cycle of lower limb angle pairs: upper leg relative to the Y~axis, lower leg relative to the upper leg, and the foot relative to the Z~axis.
\method Dikovski by Dikovski~\etal~\cite{DMG14} selects the mean during one gait cycle of step length, height, all bone lengths, then mean, standard deviation, minimum, maximum and mean difference of subsequent frames during one gait cycle of all major joint angles, and the angle between the lines of the shoulder joints and the hip joints.
\method Gavrilova by Gavrilova~\etal~\cite{APG15} chooses 20 joint relative distance signals and 16~joint relative angle signals across the whole body, compared using the DTW.
\method Jiang by Jiang~\etal~\cite{JWZS15} measures angle signals between the Y~axis and four major lower body (thigh and calf) bones. The signals are compared using the DTW.
\method Krzeszowski by Krzeszowski~\etal~\cite{KSKJW14} observes the signals of rotations of eight major bones (humerus, ulna, thigh and calf) around all three axes, the person's height and step length. These signals are compared using the DTW distance function.
\method Kumar by Kumar~\etal~\cite{NV12} extracts all joint trajectories around all three axes. Gait samples are compared by a distance function of their covariance matrices.
\method Kwolek by Kwolek~\etal~\cite{KKMJ14} processes signals of bone angles around all axes, the person's height and step length. The gait cycles are normalized to 30~frames.
\method Preis by Preis~\etal~\cite{PKWL12} takes height, length of legs, torso, both lower legs, both thighs, both upper arms, both forearms, step length and speed.
\method Sedmidubsky by Sedmidubsky~\etal~\cite{SVBZ12} concludes that only the two shoulder-hand signals are discriminatory enough to be used for recognition. These temporal data are compared using the DTW distance function.
\method Sinha by Sinha~\etal~\cite{SCB13} combines all features of Ball and Preis with mean areas during one gait cycle of upper body and lower body, then mean, standard deviation and maximum distances during one gait cycle between the centroid of the upper body polygon and the centroids of four limb polygons.
\end{itemize}

We are interested in finding an optimal feature space by maximizing its class separability, which is when gait templates are close to those of the same walker and far from those of other walkers. The method proposed in~\cite{BS16a,BS16b} learns gait features directly from joint coordinates by a modification of Fisher’s Linear Discriminant Analysis~\cite{F36} with Maximum Margin Criterion. The framework allows learning from bone rotations as well.

Let the model of a human body have $\gJ$~joints and all samples be linearly normalized to their average length~$\gT$. Labeled learning data in the sample (measurement) space $\gL{\gG}$ are in the form $\left\{\left(\gGn{n},\gLAMBDAn{n}\right)\right\}_{n=1}^{\gL{\gN}}$ where $\gGn{n}=\left[[\gGAMMAjt{1}{1}\,\cdots\,\gGAMMAjt{\gJ}{1}]^\top\,\cdots\,[\gGAMMAjt{1}{\gT}\,\cdots\,\gGAMMAjt{\gJ}{\gT}]^\top\right]^\top$ is a sample (gait cycle) in which $\gGAMMAjt{j}{t}\in\mathbb{R}^3$ are 3D spatial coordinates of a joint $j\in\left\{1,\ldots,\gJ\right\}$ at time $t\in\left\{1,\ldots,\gT\right\}$ normalized with respect to the person's position and direction. See that $\gL{\gG}$ has dimensionality $\gD=3\gJ\gT$. Learning on bone rotations is analogical. Each learning sample falls strictly into one of the learning identity classes $\left\{\gIc{c}\right\}_{c=1}^{\gC}$ determined by $\gLAMBDAn{n}$. A~class $\gIc{c}\subseteq\gL{\gG}$ has $\gNc{c}$ samples. The classes are complete and mutually exclusive. We say that learning samples $\left(\gGn{n},\gLAMBDAn{n}\right)$ and $\left(\gGn{n'},\gLAMBDAn{n'}\right)$ share a common walker if and only if they belong to the same class, i.e., $\left(\gGn{n},\gLAMBDAn{n}\right),\left(\gGn{n'},\gLAMBDAn{n'}\right)\in\gIc{c}\Leftrightarrow\gLAMBDAn{n}=\gLAMBDAn{n'}$.

Apart from Maximum Margin Criterion (MMC) we also investigated the fusion of Principal Component Analysis (PCA) with Linear Discriminant Analysis (LDA) that has been used for silhouette-based (2D) gait recognition by Su~\etal~\cite{SLC09}. Feature extraction is given by a linear transformation (feature) matrix $\gPHI\in\mathbb{R}^{\gD\times\gH{\gD}}$ from a \hbox{$\gD$-dimensional} sample space $\gG=\left\{\gGn{n}\right\}_{n=1}^{\gN}$ of not necessarily labeled gait samples to a $\gH{\gD}$-dimensional feature space $\gH{\gG}=\left\{\gGnH{n}\right\}_{n=1}^{\gN}$ of gait templates where $\gH{\gD}<\gD$ and gait samples $\gGn{n}$ are transformed into gait templates $\gGnH{n}$ by $\gGnH{n}=\gPHI^\top\gGn{n}$.

On given labeled learning data~$\gL{\gG}$, Algorithm~\ref{a1} and Algorithm~\ref{a2} are efficient ways of learning the transforms $\gPHI$ for MMC and PCA+LDA, respectively. Both algorithms~\cite{BS16a,BS16b} are of quadratic complexity with respect to the number of learning identity classes due to the singular value decomposition and eigenvalue decomposition.

\begin{algorithm}[ht]
\caption{LearnTransformationMatrixMMC$\left(\gL{\gG}\right)$}
\label{a1}
\begin{algorithmic}[1]
  \State split $\gL{\gG}=\left\{\left(\gGn{n},\gLAMBDAn{n}\right)\right\}_{n=1}^{\gL{\gN}}$ into classes $\left\{\gIc{c}\right\}_{c=1}^{\gL{\gC}}$ of $\gNc{c}=\left|\gIc{c}\right|$ samples
  \State compute overall mean $\gM=\frac{1}{\gL{\gN}}\sum_{n=1}^{\gL{\gN}}\gGn{n}$ and individual class means $\gMc{c}=\frac{1}{\gNc{c}}\sum_{n=1}^{\gNc{c}}\gGnc{n}{c}$
  \State compute $\gSIGMAb=\sum_{c=1}^{\gL{\gC}}\left(\gMc{c}-\gM\right)\left(\gMc{c}-\gM\right)^\top$
  \State compute $\gCHI=\frac{1}{\sqrt{\gL{\gN}}}\left[\left(\gGn{1}-\gM\right)\cdots\left(\gGn{\gL{\gN}}-\gM\right)\right]$
  \State compute $\gUPSILON=\left[\left(\gMc{1}-\gM\right)\cdots\left(\gMc{\gL{\gC}}-\gM\right)\right]$
  \State compute eigenvectors $\gOMEGA$ and corresponding eigenvalues $\gTHETA$ of $\gSIGMAt$ through SVD of $\gCHI$
  \State compute eigenvectors $\gXI$ of $\gTHETA^{\nicefrac{-1}{2}}\gOMEGA^\top\gSIGMAb\gOMEGA\gTHETA^{\nicefrac{-1}{2}}$ through SVD of $\gTHETA^{\nicefrac{-1}{2}}\gOMEGA^\top\gUPSILON$
  \State compute eigenvectors $\gPSI=\gOMEGA\gTHETA^{\nicefrac{-1}{2}}\gXI$
  \State compute eigenvalues $\gDELTA=\gPSI^\top\gSIGMAb\gPSI$
  \State return transform $\gPHI$ as eigenvectors in $\gPSI$ that correspond to the eigenvalues of at least $\nicefrac{1}{2}$ in $\gDELTA$
\end{algorithmic}
\end{algorithm}
\vspace{-15pt}
\begin{algorithm}[ht]
\caption{LearnTransformationMatrixPCALDA$\left(\gL{\gG}\right)$}
\label{a2}
\begin{algorithmic}[1]
  \State split $\gL{\gG}=\left\{\left(\gGn{n},\gLAMBDAn{n}\right)\right\}_{n=1}^{\gL{\gN}}$ into classes $\left\{\gIc{c}\right\}_{c=1}^{\gL{\gC}}$ of $\gNc{c}=\left|\gIc{c}\right|$ samples
  \State compute overall mean $\gM=\frac{1}{\gL{\gN}}\sum_{n=1}^{\gL{\gN}}\gGn{n}$ and individual class means $\gMc{c}=\frac{1}{\gNc{c}}\sum_{n=1}^{\gNc{c}}\gGnc{n}{c}$
  \State compute $\gSIGMAb=\sum_{c=1}^{\gL{\gC}}\left(\gMc{c}-\gM\right)\left(\gMc{c}-\gM\right)^\top$
  \State compute $\gSIGMAw=\sum_{c=1}^{\gL{\gC}}\frac{1}{\gNc{c}}\sum_{n=1}^{\gNc{c}}\left(\gGnc{n}{c}-\gMc{c}\right)\left(\gGnc{n}{c}-\gMc{c}\right)^\top$
  \State compute eigenvectors $\gPHI_\mathrm{PCA}$ of $\gSIGMAt=\gSIGMAb+\gSIGMAw$ that correspond to $\gO{\gD}=\gL{\gC}$ largest eigenvalues
  \State compute eigenvectors $\gPHI_\mathrm{LDA}$ of $(\gPHI_\mathrm{PCA}^\top\gSIGMAw\gPHI_\mathrm{PCA})^{-1}(\gPHI_\mathrm{PCA}^\top\gSIGMAb\gPHI_\mathrm{PCA})$
  \State return transform $\gPHI=\gPHI_\mathrm{PCA}\gPHI_\mathrm{LDA}$
\end{algorithmic}
\end{algorithm}

In addition to the gait features extraction methods of our fellow researchers, we implemented our own methods as described below. Depending on whether the raw data are in the form of bone rotations or joint coordinates, the methods are referred to with BR or JC subscripts, respectively.
\begin{itemize}
\method \_MMC learns gait features by MMC (Algorithm~\ref{a1}) and the gait templates are compared by the Mahalanobis distance~\cite{BS16a,BS16b}.
\method \_PCALDA learns gait features by PCA+LDA (Algorithm~\ref{a2}) and the gait templates are also compared by the Mahalanobis distance~\cite{BS16a,BS16b}.
\method \_Random has no features and classification is performed by picking a random identity that is present in the gallery.
\method \textbf{\_Raw} takes all raw data. The template vector, normalized to the average of $\gT=150$ frames, results in a large feature space dimensionality $\gH{\gD}=\gD=3\gJ\gT=13{,}950$, which is why the raw data cannot be directly used for recognition on large databases.
\end{itemize}

\section{Evaluation}
\label{eval}

\begin{figure}[ht]
\centering
\includegraphics[width=0.8\textwidth]{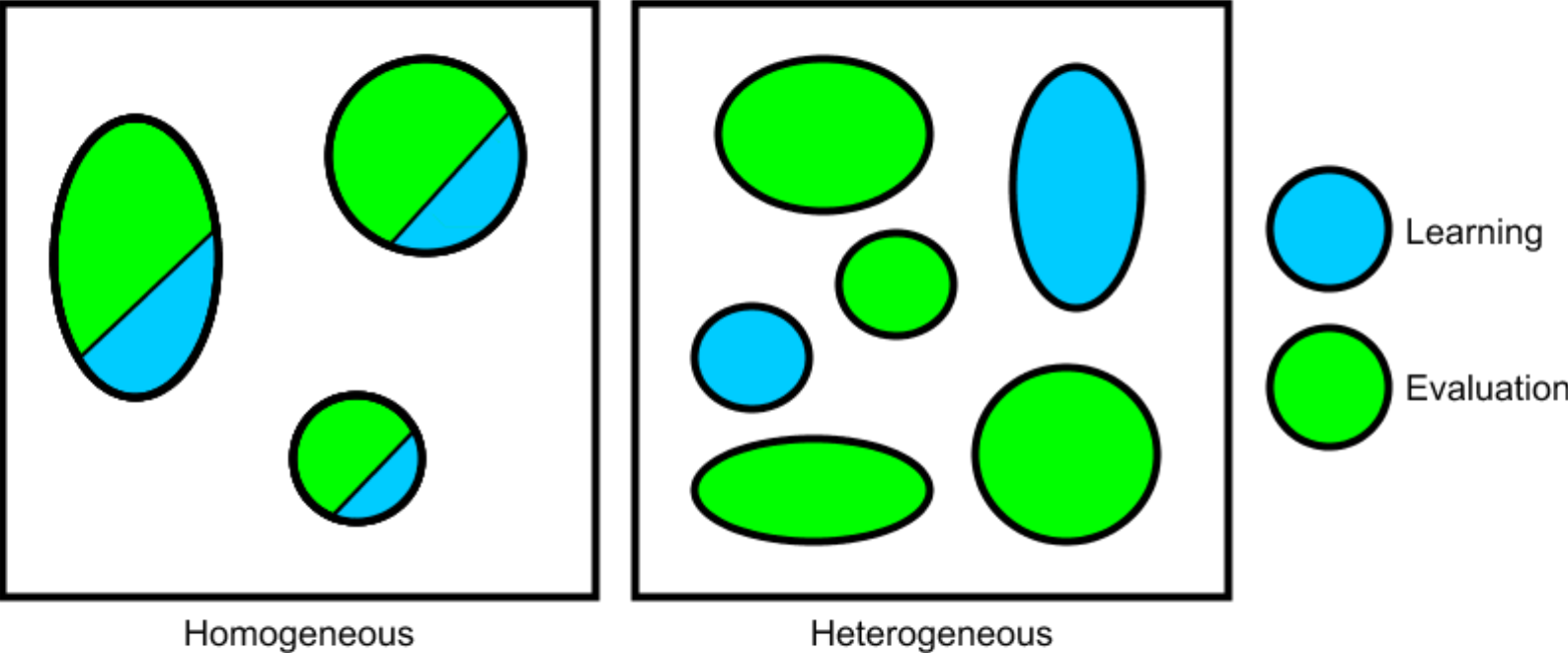}
\caption{Data separation for the homogeneous setup of $\gL{\gC}=\gE{\gC}=3$ learning-and-evaluation classes (left) and for the heterogeneous setup of~$\gL{\gC}=2$ learning classes and $\gE{\gC}=4$ evaluation classes (right). The black square represents a database and the ellipses are the identity classes.}
\label{f2}
\end{figure}

Learning data $\gL{\gG}=\left\{\left(\gGn{n},\gLAMBDAn{n}\right)\right\}_{n=1}^{\gL{\gN}}$ of $\gL{\gC}$ identities and evaluation data $\gE{\gG}=\left\{\left(\gGn{n},\gLAMBDAn{n}\right)\right\}_{n=1}^{\gE{\gN}}$ of $\gE{\gC}$ identity classes have to be disjunct at all times. In the following, we introduce two setups of data separation: homogeneous and heterogeneous. The homogeneous setup learns the transformation matrix on $\nicefrac{1}{3}$ samples of $\gL{\gC}$ identities and is evaluated on templates derived from the other $\nicefrac{2}{3}$ samples of the same $\gE{\gC}=\gL{\gC}$ identities. The heterogeneous setup learns the transform on all samples in $\gL{\gC}$ identities and is evaluated on all templates derived from other $\gE{\gC}$ identities. An abstraction of this concept is depicted in Figure~\ref{f2}. Note that unlike in the homogeneous setup, no walker identity is ever used for both learning and evaluation at the same time in the heterogeneous setup.

The homogeneous setup is parametrized by a single number $\gL{\gC}=\gE{\gC}$ of learning-and-evaluation identity classes, whereas the heterogeneous setup has the form $\left(\gL{\gC},\gE{\gC}\right)$ specifying how many learning and how many evaluation identity classes are randomly selected from the database. The evaluation of each setup is repeated 3~times, selecting new random $\gL{\gC}$ and $\gE{\gC}$ identity classes each time and reporting the average result.

\begin{figure}[ht]
\centering
\includegraphics[width=0.8\textwidth]{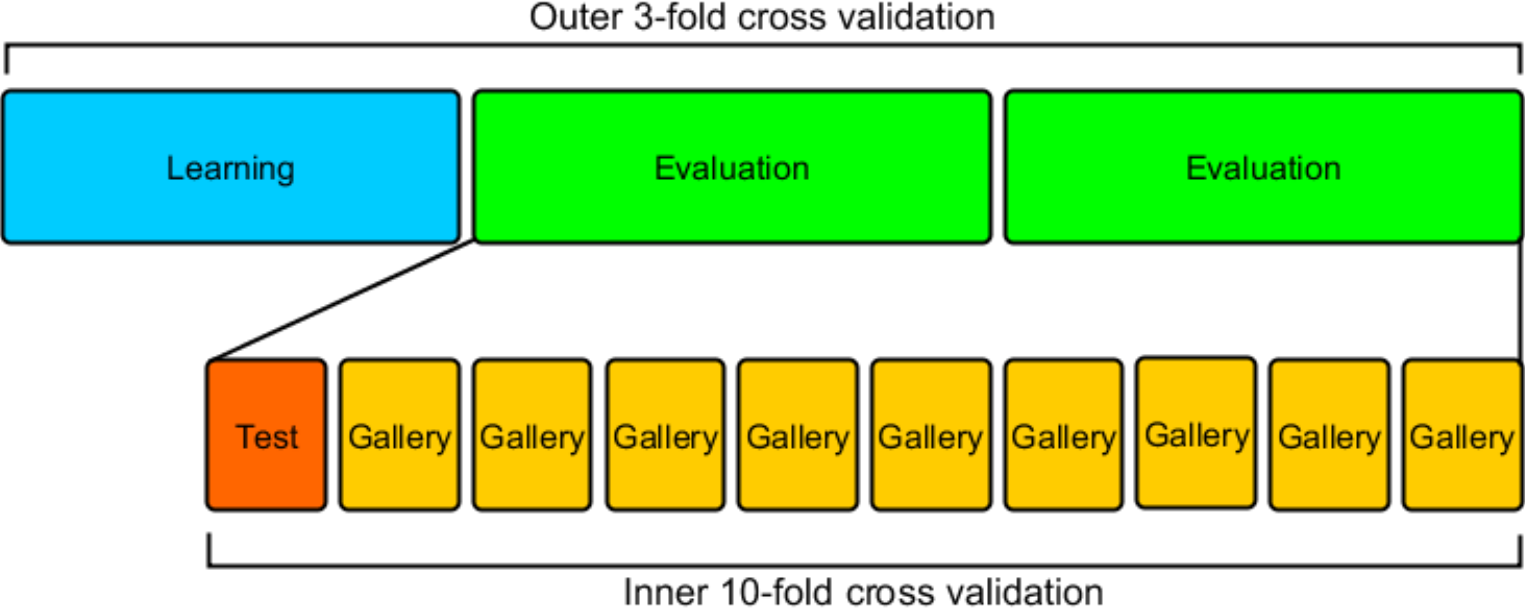}
\caption{Nested cross-validation.}
\label{f7}
\end{figure}

In the homogeneous setup, all results are estimated with nested cross-validation (see Figure~\ref{f7}) that involves the outer 3-fold cross-validation loop where templates in one fold are used for learning the features, while templates in the remaining two folds are used for evaluations. In the heterogeneous setup, the learning and evaluation parts are selected at random based on the given $\gL{\gC}$ and $\gE{\gC}$, respectively. For both setups, this model is frozen and ready to be evaluated for class separability coefficients. Evaluation of rank-based classifier performance metrics advances to the inner 10-fold cross-validation loop taking one dis-labeled fold as a testing set and the other nine labeled folds as gallery. Test templates are classified by the winner-takes-all strategy, in which a test template $\gGnH{}^{\mathrm{test}}$ gets assigned with the label $\gLAMBDAn{\argmin_i\gDELTAccH{\gGnH{}^{\mathrm{test}}}{\gGnH{i}^{\mathrm{gallery}}}}$ of the gallery's closest identity class.

Correct Classification Rate (CCR) is often perceived as the ultimate qualitative measure, however, if a method has a low CCR, we cannot directly say if the system is failing because of bad features or a bad classifier. It is more explanatory to provide an evaluation in terms of class separability of the feature space. The class separability measures give an estimate on the recognition potential of the extracted features and do not reflect an eventual combination with an unsuitable classifier:
\begin{description}[style=unboxed,leftmargin=1em]
\def\ditem #1 (#2){\item[\textbullet\enskip\normalfont{\textit{#1:}}\enskip\textbf{#2}]}
\ditem Davies-Bouldin Index (DBI)
\begin{equation}
\mathrm{DBI}=\frac{1}{\gE{\gC}}\sum_{c=1}^{\gE{\gC}}\max\limits_{1 \leq c' \leq \gE{\gC},\,c' \neq c}\frac{\sigma_c+\sigma_{c'}}{\gDELTAccH{\gMcH{c}}{\gMcH{c'}}}
\end{equation}
where $\sigma_c=\frac{1}{\gNc{c}}\sum_{n=1}^{\gNc{c}}\gDELTAccH{\gGnH{n}}{\gMcH{c}}$ is the average distance of all elements in identity class $\gIc{c}$ to its centroid, and analogically for $\sigma_{c'}$. Templates of low intra-class distances and of high inter-class distances have a low DBI.
\ditem Dunn Index (DI)
\begin{equation}
\mathrm{DI}=\frac{\min\limits_{1 \leq c<c' \leq \gE{\gC}}\gDELTAccH{\gMcH{c}}{\gMcH{c'}}}{\max\limits_{1 \leq c \leq \gE{\gC}}\sigma_c}
\end{equation}
with $\sigma_c$ from the above DBI. Since this criterion seeks classes with high intra-class similarity and low inter-class similarity, a high DI is more desirable.
\ditem Silhouette Coefficient (SC)
\begin{equation}
\mathrm{SC}=\frac{1}{\gE{\gN}}\sum_{n=1}^{\gE{\gN}}\frac{b(\gGnH{n})-a(\gGnH{n})}{\max\left\{a\left(\gGnH{n}),b(\gGnH{n}\right)\right\}}
\end{equation}
where $a(\gGnH{n})=\frac{1}{\gNc{c}}\sum_{n'=1}^{\gNc{c}}\gDELTAccH{\gGnH{n}}{\gGnH{n'}}$ is the average distance from $\gGnH{n}$ to other samples within the same identity class and $b(\gGnH{n})=\min\limits_{1 \leq c' \leq \gE{\gC},\,c' \neq c}\frac{1}{\gNc{c'}}\sum_{n'=1}^{\gNc{c'}}\gDELTAccH{\gGnH{n}}{\gGnH{n'}}$ is the average distance of $\gGnH{n}$ to the samples in the closest class. It is clear that $-1 \leq \mathrm{SC} \leq 1$ and SC close to one means that classes are appropriately separated.
\ditem Fisher's Discriminant Ratio (FDR)
\begin{equation}
\mathrm{FDR}=\frac{\frac{1}{\gC}\sum_{c=1}^{\gE{\gC}}\gDELTAccH{\gMcH{c}}{\gH{\gM}}}{\frac{1}{\gE{\gN}}\sum_{c=1}^{\gE{\gC}}\sum_{n=1}^{\gNc{c}}\gDELTAccH{\gGnH{n}}{\gMcH{c}}}.
\end{equation}
High FDR is preferred for classes of low intra-class sparsity and high inter-class sparsity.
\end{description}

Apart from analyzing the distribution of templates in the feature space, it is schematic to combine the features with a rank-based classifier and to evaluate the system based on distance distribution with respect to a probe. For obtaining a more applied performance evaluation, we evaluate:
\begin{description}[style=unboxed,leftmargin=1em]
\def\ditem #1 (#2){\item[\textbullet\enskip\normalfont{\textit{#1:}}\enskip\textbf{#2}]~\newline}
\ditem Cumulative Match Characteristic (CMC) 
Sequence of Rank-$k$ (for $k$ on X axis from~1 up to $\gE{\gC}$) recognition rates (Y~axis) for measuring ranking capabilities of a recognition method. Its headline Rank-1 is the well-known \textbf{CCR}.
\ditem False Accept Rate vs.\@ False Reject Rate (FAR/FRR)
Two sequences of the error rates (Y~axis) as functions of discrimination threshold (X~axis). Each method has a value~$e$ of this threshold giving Equal Error Rate (\textbf{EER}=FAR=FRR).
\ditem Receiver Operating Characteristic (ROC)
Sequence of True Accept Rate (\textbf{TAR}) and False Accept Rate (\textbf{FAR}) with a varied discrimination threshold. For a given threshold the system signals both TAR (Y~axis) and FAR (X~axis). Area Under Curve (\textbf{AUC}) is computed as the integral of the ROC curve.
\ditem Recall vs.\@ Precision (RCL/PCN)
Sequence of rates with a varied discrimination threshold. For a given threshold the system signalizes both RCL (X~axis) and PCN (Y~axis). The value of Mean Average Precision (\textbf{MAP}) is computed as the area under RCL/PCN curve.
\end{description}

These measures reflect how well the feature is class-separated and how much it takes to confuse the identities of two people. They do not, in fact, provide complementary information, however, a quality evaluation framework should be able to evaluate the most popular measures. Each measure is evaluated in the context of a particular application. For example, a hotel lobby authentication system could use a high Rank-3 at the CMC, while a city-level person tracking system is likely to need the ROC curve leaning towards the upper left corner.

\section{Reproducing the Experiments}
\label{repro}

This section provides a description of the framework we implemented and the database we extracted. With this manual, a reader should be able to reproduce the evaluation and to use the implementation for recognizing people. All source codes including (1)~database extraction drive, (2)~implementations of the proposed and all relevant methods, (3)~classifier learning and classification mechanisms and (4)~evaluation mechanism and metrics, are available at our departmental Git repository~\cite{WWW1}. The original CMU MoCap database and extracted databases are available online at our research group web page~\cite{WWW2}.

\texttt{Executor.java} is the main execution class. Set all parameters of evaluation and  file locations and \texttt{distanceThreshold}, then select which actions to perform and finally, select the evaluation setups. The class contains the \texttt{main(String[] args)} method. It contains four methods to select for execution:
\begin{description}
\item[\texttt{extractDatabase()}] for extracting an experimental database from the original CMU MoCap database~–-~this is also available for download at our page as \texttt{original.zip}. To run this method, unzip to get the files \texttt{gaitcycle.amc} (exemplary gait cycle) and \texttt{skeleton.asf} (prototypical skeleton) and the directory \texttt{amcOriginal} (original AMC files). Extraction begins with normalization with respect to a person’s position and walk direction as provided in the \texttt{normalized.zip} file. Clean gait cycles are then filtered out by the distance threshold (see last paragraph of Section~\ref{db}) that numerically expresses how much extracted motions resemble gait cycles, that is, the lower the distance threshold, the fewer and cleaner the gait cycles. Set a value for \texttt{distanceThreshold} to produce a folder of the extracted database. Evaluations in~\cite{BS16a,BS16b} are set with 302.0, extracting a database of 54~identities and 3{,}843~gait cycles. A higher distance threshold will qualify some non-gait motions.
\item[\texttt{learnClassifiers()}] for learning classifiers of all implemented methods on a sub-database determined by the distance threshold. Set a value for \texttt{distanceThreshold} (such as 302.0) and provide the corresponding directory of the learning database (such as \texttt{amc302.0} in \texttt{extracted-302.0.zip}) and the learned classifiers appear in the \texttt{classifiers} folder.
\item[\texttt{performClassification()}] for performing a classification of a custom probe/query on a custom gallery with a custom classifier. Set file locations for the classifier file \texttt{customClassifier}, the probe gait cycle \texttt{customQueryFileAMC} and the gallery directory \texttt{customGalleryDirectory}. Results are printed on the standard output.
\item[\texttt{evaluateMethods()}] for evaluating the implemented methods in homogeneous and heterogeneous setups. To skip database extraction, one could supply a provided extracted database (such as \texttt{amc302.0} in \texttt{extracted-302.0.zip}) and a skeleton file (such as \texttt{skeleton.asf} in any extracted database ZIP file). Our page provides additional databases categorised according to various values of \texttt{distanceThreshold}. The results are set to be printed on the standard output but we suggest to redirect it to a CSV file. Results of individual evaluation attempts vary slightly as different learning, testing and gallery sets are randomly selected upon each attempt.
\end{description}

Compile to obtain \texttt{Gait.jar}. The main project location should also contain the \texttt{lib} directory and all necessary files and directories depending on which actions are to be executed. Run command \texttt{\$ java -jar Gait.jar > output.csv}.

The output file (see structure in Table~\ref{t1}) contains the performance metrics as specified in Section~\ref{eval} and the information about average distance computation time (\textbf{DCT}) in milliseconds and average template dimensionality (\textbf{TD}). The evaluation results are in the form of one value per coefficient (see results in Table~\ref{t2}) and the sequences CMC, FAR/FRR, ROC=TAR/FAR and RCL/PCN. The CMC sequence contains $\gE{\gC}$ values, one for each $k$ in the Rank-$k$ recognition rate. The other three pairs of sequences are normalized to 30~values by \texttt{method.setFineness(30)}. The FAR/FRR sequences of all methods are normalized to the discrimination threshold with respect to the first value of FAR=0 and FRR=1, and to the middle value that represents EER where all sequences cross. The ROC sequences are normalized with respect to the first value of TAR=FAR=0 and to the last value of TAR=FAR=1. Finally, the RCL/PCN sequences are normalized with respect to the first value of RCL=0 and to the last value of RCL=1.

\begin{table}[ht]
\vspace{-10pt}
\caption{Structure of the output file.}
\label{t1}
\vspace{-5pt}
\centering
\begin{tabular}{l}
\toprule[1pt]
\{method name\}, \{distance threshold\}\\
\tt DBI DI SC FDR CCR EER AUC MAP DCT TD\\
\{1 line -- one value for each coefficient\}\\
\tt CMC\\
\{$\gE{\gC}$ lines -- the CMC sequence\}\\
\tt FAR FRR TAR FAR RCL PCN\\
\{30 lines -- all six sequences\}\\
\bottomrule[1pt]
\end{tabular}
\vspace{20pt}
\caption{First line results of all 20 implemented methods on the 302.0 database.}
\label{t2}
\vspace{-5pt}
\centering\underworks\tabcolsep3.3pt
\begin{tabular}{r|rlll|llll|rr}
\toprule[1pt]
& \multicolumn{4}{c|}{class separability coefficients} & \multicolumn{4}{c|}{classification based metrics} & \multicolumn{2}{c}{scalability} \\
method	& DBI_	& _DI	& _SC	& FDR	& CCR	& EER	& AUC	& MAP	& _DCT	& TD \\
\midrule[0.4pt]
Ahmed	& 216.2	& 0.842	& $-$0.246	& 0.954	& 0.657	& 0.38	& 0.659	& 0.165	& \textbf{<1}	& 24 \\
Ali	& 501.5	& 0.26	& $-$0.463	& 1.175	& 0.225	& 0.384	& 0.679	& 0.111	& \textbf{<1}	& 2 \\
Andersson	& 142.3	& 1.297	& $-$0.102	& 1.127	& 0.84	& 0.343	& 0.715	& 0.251	& \textbf{<1}	& 68 \\
Ball	& 161\hphantom{.}_	& 1.458	& $-$0.163	& 1.117	& 0.75	& 0.346	& 0.711	& 0.231	& \textbf{<1}	& 18 \\
Dikovski	& 144.5	& 1.817	& $-$0.135	& \textbf{1.227}	& 0.881	& 0.363	& 0.695	& 0.254	& \textbf{<1}	& 71 \\
Gavrilova	& 185.8	& 1.708	& $-$0.164	& 0.77	& 0.891	& 0.374	& 0.677	& 0.254	& 45	& 5,254 \\
Jiang	& 206.6	& 1.802	& $-$0.249	& 0.85	& 0.811	& 0.395	& 0.657	& 0.242	& 8	& 584 \\
Krzeszowski	& 154.1	& 1.982	& $-$0.147	& 0.874	& 0.915	& 0.392	& 0.662	& 0.275	& 35	& 3,795 \\
Kumar	& \textbf{118.6}	& 1.618	& $-$0.086	& 1.09	& 0.801	& 0.459	& 0.631	& 0.217	& 8	& 13,950 \\
Kwolek	& 150.9	& 1.348	& $-$0.084	& 1.175	& 0.896	& 0.358	& 0.723	& 0.323	& \textbf{<1}	& 660 \\
Preis	& 1,980.6	& 0.055	& $-$0.512	& 1.067	& 0.143	& 0.401	& 0.626	& 0.067	& \textbf{<1}	& 13 \\
Sedmidubsky	& 398.1	& 1.35	& $-$0.425	& 0.811	& 0.543	& 0.388	& 0.657	& 0.149	& \textbf{<1}	& 292 \\
Sinha	& 214.8	& 1.112	& $-$0.215	& 1.101	& 0.674	& 0.356	& 0.697	& 0.191	& \textbf{<1}	& 45 \\
\midrule[0.4pt]
\_MMC\sub{BR}	& 154.2	& 1.638	& \textbf{\hphantom{$-$}0.062}	& 1.173	& 0.925	& \textbf{0.297}	& \textbf{0.748}	& 0.353	& \textbf{<1}	& 53 \\
\_MMC\sub{JC}	& 130.3	& 1.891	& \hphantom{$-$}0.051	& 1.106	& 0.918	& 0.378	& 0.721	& 0.315	& \textbf{<1}	& 51 \\
\_PCALDA\sub{BR}	& 182\hphantom{.}_	& 1.596	& $-$0.015	& 0.984	& 0.918	& 0.361	& 0.695	& 0.276	& \textbf{<1}	& 54 \\
\_PCALDA\sub{JC}	& 174.4	& 1.309	& $-$0.091	& 0.827	& 0.863	& 0.44	& 0.643	& 0.201	& \textbf{<1}	& 54 \\
\_Random	&	&	&	&	& 0.042	&	&	&	&	& \textbf{0} \\
\_Raw\sub{BR}	& 163.7	& \textbf{2.092}	& \hphantom{$-$}0.011	& 0.948	& \textbf{0.966}	& 0.315	& 0.743	& \textbf{0.358}	& 70	& 8,229 \\
\_Raw\sub{JC}	& 155.1	& 1.954	& $-$0.12	& 0.897	& 0.926	& 0.377	& 0.679	& 0.283	& 161	& 13,574 \\
\bottomrule[1pt]
\end{tabular}
\end{table}

To reproduce the experiments in Table~\ref{t2}, follow the instructions in the README file at~\cite{WWW1} in the \texttt{reproduce} folder. Please note that some methods are slow even on a~leading edge hardware. Learning and evaluation times in Table~\ref{t3} were measured on a~computer with Intel\textregistered\ Xeon\textregistered\ CPU E5-2650 v2 @ 2.60GHz and \unit[256]{GB}~RAM.

\begin{table}[ht]
\vspace{-10pt}
\caption{Evaluation times of the methods in Table~\ref{t2}. Units: \unit{s}~seconds, \unit{m}~minutes, \unit{h}~hours, \unit{d}~days.}
\label{t3}
\vspace{-5pt}
\centering\underworks\tabcolsep4.2pt
\begin{tabular}{rl|rl|rl|rl}
\toprule[1pt]
method	& time	&	method	& time	&	method	&	time	&	method	& time \\
\midrule[0.4pt]
Ahmed	& \unit[48.6]{m}	&	Gavrilova	& \unit[10.3]{d}	&	Preis	& \unit[48.7]{m}	&	\_PCALDA\sub{BR}	& \unit[\hphantom{0}4.7]{h} \\
Ali	& \unit[40.9]{m}	&	Jiang	& \unit[\hphantom{0}1.9]{d}	&	Sedmidubsky	& \unit[\hphantom{0}1.4]{d}	&	\_PCALDA\sub{JC}	& \unit[10.9]{h} \\
Andersson	& \unit[45.7]{m}	&	Krzeszowski	& \unit[\hphantom{0}8.1]{d}	&	Sinha	& \unit[49.6]{m}	&	\_Random	& \unit[27.9]{m} \\
Ball	& \unit[48.5]{m} 	&	Kumar	& \unit[\hphantom{0}1.8]{d}	&	\_MMC\sub{BR}	& \unit[\hphantom{0}2.6]{h}	&	\_Raw\sub{BR}	& \unit[16.1]{d} \\
Dikovski	& \unit[50.7]{m}	&	Kwolek	& \unit[\hphantom{0}1.1]{h}	&	\_MMC\sub{JC}	& \unit[\hphantom{0}3.0]{h}	&	\_Raw\sub{JC}	& \unit[36.7]{d} \\
\bottomrule[1pt]
\end{tabular}
\vspace{-12pt}
\end{table}

The goal of the MMC-based learning is to find a linear discriminant that maximizes the misclassification margin. This optimization technique appears to be more effective than designing geometric gait features. Table~\ref{t2} indicates the best results for the MMC on bone rotational data: highest SC, EER and AUC, and competitive DBI, DI, FDR, CCR and MAP. In terms of the Correct Classification Rate metric, our MMC method was only outperformed by the Raw method, which is implemented here as a form of baseline. We interpret the high scores as a sign of robustness.

Apart from the performance merits, the MMC method is also efficient: relatively low-dimensional templates and Mahalanobis distance ensure fast distance computations and thus contribute to high scalability. Note that even if the Raw method has some of the best results, it can hardly be used in practice due to its extreme consumption of time and space resources. On the other hand, Random has no features but cannot be considered a~serious recognition method. To illustrate the evaluation time, calculating the distance matrix (a matrix of distances between all evaluation templates) took a couple minutes for the MMC method, almost nothing for the Random method, and more than two weeks for the Raw method. To conclude, the MMC method on bone rotational data appears to be an optimal trade-off between effectiveness and efficiency, and thus the new state-of-the-art in feature extraction for MoCap-based gait recognition.

\section{Summary and Future Work}
\label{summ}

As our contribution to reproducible research, we have provided implementation details and source codes~\cite{WWW1} of our evaluation framework for gait recognition~\cite{BS16a,BS16b}. The software implements the proposed method as well as all related methods. We include the evaluation database~\cite{WWW2} together with source codes for its extraction from the general CMU MoCap database. We also attach the description and portable software for evaluating class separability coefficients of extracted features and classifier performance metrics. Finally, we provide documentation and installation instructions for easy and straightforward reproducibility of the experiments.

As demonstrated by outperforming other methods in four class separability coefficients and four classification metrics, the proposed features learning mechanism has a~strong potential in gait recognition applications. Even though we believe that MMC is the most suitable criterion for optimizing gait features, we continue to research further potential optimality criteria and machine learning approaches.

We hope that the evaluation framework and database presented here will contribute to smooth development and evaluation of further novel MoCap-based gait recognition methods. All used data and source codes have been made available~\cite{WWW1,WWW2} under the Creative Commons Attribution license (CC-BY) for database and the Apache~2.0 license for software, which grant free use and allow for experimental evaluation. We encourage all readers and developers of MoCap-based gait recognition methods to contribute to the framework with new algorithms, data and improvements.

\vspace{-1pt}
\subsubsection*{Acknowledgments}
The authors thank to the anonymous reviewers and editor for their detailed commentary and suggestions. The data used in this project was created with funding from NSF EIA-0196217 and was obtained from \url{http://mocap.cs.cmu.edu}~\cite{CMU03}.

\bibliographystyle{splncs03}
\bibliography{ref}
\end{document}